\documentclass[letterpaper, 10pt, conference]{ieeeconf}      

\IEEEoverridecommandlockouts                              

\usepackage{graphicx} 

\usepackage{epsfig} 
\usepackage{epstopdf}
\usepackage{tikz}
\usepackage{mathptmx} 
\usepackage{times} 
\usepackage{amsmath} 
\usepackage{amssymb}  

\usepackage{algorithm}
\usepackage{algorithmic}

\usepackage{tablefootnote}
\usepackage{color}
\usepackage{amsfonts}
\usepackage[noadjust]{cite}
\usepackage{subfigure}
\usepackage{threeparttable}
\usepackage[framed,numbered,autolinebreaks,useliterate]{mcode}
\usepackage{soul}
\usepackage[normalem]{ulem}
\usepackage{pgf}
\usepackage{tikz}
\usetikzlibrary{arrows,automata}

\usepackage{booktabs}

\usepackage[a-1b]{pdfx}

\usepackage{hyperref} 

\usepackage{todonotes}

\title{\LARGE \bf Robust Generalized Proportional Integral Control for Trajectory Tracking of Soft Actuators in a Pediatric Wearable Assistive Device} 

\author{Caio Mucchiani,$^{1}$ Zhichao Liu,$^{1}$ Ipsita Sahin,$^{2}$ Elena Kokkoni,$^{2}$ Konstantinos Karydis$^{1}$
\thanks{$^{1}$ Dept. of Electrical and Computer Eng. and $^{2}$ Dept. of Bioengineering; Univ. of California, Riverside, 900 University Avenue, Riverside, CA 92521, USA. Email: {\tt\footnotesize\{caiocesr, zliu157, isahi001, elenak, karydis\}@ucr.edu}. 
We gratefully acknowledge the support of NSF \# CMMI-2133084 and ARL \# W911NF-18-1-0266. 
Any opinions, findings, and conclusions or recommendations expressed in this material are those of the authors and do not necessarily reflect the views of the funding agencies.
}}

\begin{document}
\maketitle

\begin{abstract}
Soft robotics hold promise in the development of safe yet powered assistive wearable devices for infants. Key to this is the development of closed-loop controllers that can help regulate pneumatic pressure in the device's actuators in an effort to induce controlled motion at the user's limbs and be able to track different types of trajectories. This work develops a controller for soft pneumatic actuators aimed to power a pediatric soft wearable robotic device prototype for upper extremity motion assistance. The controller tracks desired trajectories for a system of soft pneumatic actuators supporting two-degree-of-freedom shoulder joint motion on an infant-sized engineered mannequin. The degrees of freedom assisted by the actuators are equivalent to shoulder motion (abduction/adduction and flexion/extension). Embedded inertial measurement unit sensors provide real-time joint feedback. Experimental data from performing reaching tasks using the engineered mannequin are obtained and compared against ground truth to evaluate the performance of the developed controller. Results reveal the proposed controller leads to accurate trajectory tracking performance across a variety of shoulder joint motions. 
\end{abstract}



%



\section{Introduction}

Soft robotics technology has been increasingly integrated into wearable assistive devices (e.g.,~\cite{yumbla2021human}) owing to the former's inherent flexibility and adaptability, softness, and lower profile compared to rigid-only devices~\cite{o2017soft}. 
A major portion of previous work in this area include soft wearable assistive devices for upper extremity assistance and rehabilitation of adult populations (notable examples are~\cite{proietti2021sensing, o2020inflatable, park2017development}).
Despite a range of assistive devices proposed for these populations, upper extremity assistive devices for very young pediatric populations (i.e. $<$2 years of age) are limited~\cite{arnold2020exploring, christy2016technology} and mainly passive~\cite{lobo2016playskin, babik2016feasibility}.
The present work specifically focuses on the development of a soft robotic wearable device for upper extremity assistance in infants.


Several considerations should be taken into account when developing an assistive device for a young pediatric population; such as their anthropometric, movement, and learning characteristics~\cite{wininger2017geek}. 
For example, infants' kinematic parameters of motion (such as velocity profile of the hand) differ from those of older populations~\cite{konczak1997development,morange2019visual}. 
Soft wearable assistive devices, in particular, can afford a wide variety of methods for actuation, sensing and control, which in many cases are intertwined with each other and can be adjusted to meet the characteristics of the infant population. 
Actuation methods can utilize 3D-printing~\cite{yap2016high,schaffner20183d,hoang2021pneumatic}, casting~\cite{li2020high,kokkoni2020development,liu2022safely} or fabric~\cite{yap2017fully,kim2021compact,nassour2020high,fu2022textiles,sahin2022bidirectional} to improve strength~\cite{simpson2020upper,o2017soft} and minimize fatigue~\cite{nassour2020high,o2020inflatable}. 
Several of the current actuator designs employed in wearable devices are pneumatic-based (as in this work too), in an effort to facilitate motion while providing comfort and safety with minimal constraints on the arms~\cite{diteesawat2022soft}. 

Given a higher-level controller (e.g., admittance force control) 
yielding desired trajectories 
(e.g., as seen in other soft wearable assistive devices~\cite{zhou2021human}), 
a low-level controller operating at the pneumatic-system level needs to ensure the desired trajectory is achieved. 
%
In our past work we have developed feedforward~\cite{kokkoni2020development,liu2021position} as well as setpoint~\cite{mucchiani2022closed} and learning-based feedback~\cite{shi2022online} controllers for soft pneumatic actuators used in various iterations of our wearable device prototypes~\cite{kokkoni2020development,sahin2022bidirectional}. Feedback controllers can help make the device more responsive, while tracking desired setpoints is a key function on its own, e.g., to counter gravity at the shoulder joint~\cite{proietti2021sensing}. However, to make soft wearable assistive devices for infant reaching more capable, it is important to be able to track complete trajectories too. 

\begin{figure}[!t]
   \vspace{6pt}
   \centering
     \includegraphics[width=0.99\columnwidth]{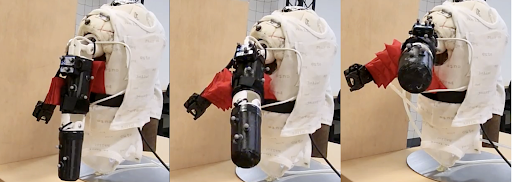}
     \vspace{-18pt}
      \caption{Snapshots of a combined joint angle trajectory motion of the assisted shoulder. The accompanying video offers more visual information and contains footage regarding the experiments conducted herein.}
      \label{fig:intro}
      \vspace{-6pt}
\end{figure}

In this work, we develop a low-level feedback robust generalized proportional integral (GPI) controller to track desired trajectories for a system of soft pneumatic actuators supporting shoulder joint motion on an infant-sized engineered mannequin (Fig.~\ref{fig:intro}). A GPI low-level controller is suitable in the context of this work as it can track trajectories while requiring less control effort compared to standard-of-practice PID controllers~\cite{blanco2022robust}. Further, it only requires knowledge of the mannequin's (and eventually user's) arm joint angles which can in principle reduce the computational load. GPI controllers have been used in conceptually-related works on exoskeletons for rehabilitation purposes of both upper~\cite{blanco2022robust} and lower~\cite{azcaray2018robust} extremities. The former considered the same controllable degrees of freedom (DoFs) as we do herein for the shoulder joint; however, rigid, rather than soft actuators were used, and the study did not specifically aim at pediatric populations. 
%
%
Our approach critically focuses on a system of soft pneumatic actuators tailored for pediatric populations. Our engineered mannequin is designed and fabricated in-house based on average 12-month-old infant anthropometrics~\cite{Fryar2021,edmond2020normal}. 
Experimental testing and evaluation of the performance of the proposed low-level controller off-body using a mannequin is a critical and necessary step prior to being able to test with human subjects.

Our overall system contributes to on-board sensor-based pneumatic feedback control for soft robotics with application to wearable assistive devices. We consider two types of fabric-based soft pneumatic actuators for supporting up to two-DoF-equivalent shoulder motion (abduction/adduction--AB/AD and flexion/extension--F/E). 
Joint angle variations of the mannequin in response to actuator inflation/deflation are estimated in real time via proprioceptive feedback from inertial measurement units (IMUs). Then, given a desired trajectory in the joint space (relevant to the mannequin), our developed GPI controller regulates the pneumatic actuation Pulse-Width-Modulation (PWM) values to track the desired trajectory. 
Our proposed controller is shown to track correctly interpolated trajectories for both AB/AD and F/E shoulder motion separately, as well as simultaneously. In addition, to evaluate system and controller robustness under expected motion primitives common to infants, periodic trajectories in the form of harmonic signals and teach-and-repeat trajectories are also tested and validated experimentally.

Succinctly, the paper's contributions are: 
\begin{itemize}
    \item Development of a GPI controller for joint angle trajectory tracking of soft pneumatic actuators with embedded proprioceptive sensing.
    \item Experimentation with a system of two soft actuators supporting shoulder mobility, performing trajectories aimed at motion assistance in infant reaching tasks.
\end{itemize}

To the best of our knowledge, no previous work in infant reaching has considered two DoF shoulder assistance with general motion of the limbs, neither using soft actuators (specifically of the bellow type) nor applying GPI control methods to track desired arm trajectories. 

\section{System Description}
\label{sec:problem}

\subsection{Hardware Design and Integration}
\begin{figure}[!t]
\vspace{6pt}
\centering
\includegraphics[width=0.99\linewidth]{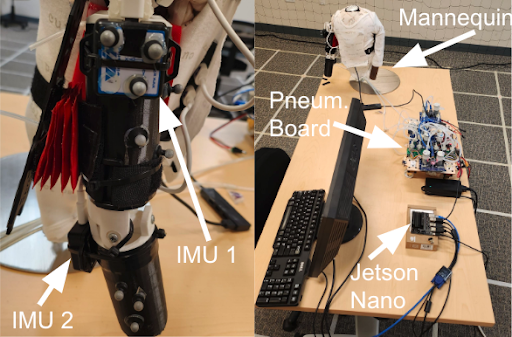}
\vspace{-18pt}
\caption{Hardware with embedded IMUs (left) and experimental setup (right). Different key components are highlighted. The two IMUs are placed so that the supported shoulder abduction/adduction (AB/AD) and flexion/extension (F/E) joint angles can be readily computed.}
\label{fig:comp}
\vspace{0pt}
\end{figure}

The overall hardware system we have designed in this work comprises the actuation subsystem (soft actuators and pneumatic control board), the mannequin where the actuators are mounted on, and the sensing and computing units required for control design in practice. The different components are highlighted in Fig.~\ref{fig:comp}. 

\begin{figure}[!t]
\vspace{-3pt}
\includegraphics[width=\linewidth]{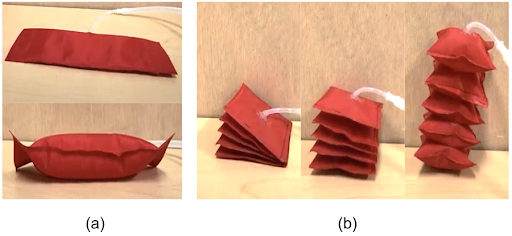}
\vspace{-21pt}
\caption{We employ two types of fabric-based soft pneumatic actuators designed in-house, with the aim to support (a) shoulder abduction/adduction and (b) shoulder flexion/extension.} \label{fig:act}
\vspace{0pt}
\label{fig:sac}
\end{figure}

\begin{figure}[!t]
\vspace{-3pt}
\centering 
\includegraphics[width=\columnwidth]{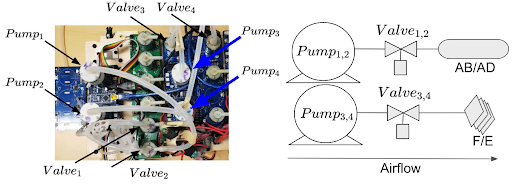}
\vspace{-25pt}
\caption{Left: The pneumatic control board designed in-house to operate the system's soft actuators.
Right: Pumps and solenoid valves diagram.}
\label{fig:pneu}
\vspace{-12pt}
\end{figure}

The system features two soft pneumatic actuators made of flexible thermoplastic polyurethane (TPU) coated nylon fabric (Fig.~\ref{fig:sac}) that provides support for two DoF motion assistance of the shoulder (AB/AD and F/E).\footnote{~Here we consider analysis of single-arm motion. The exact same actuation system can be readily replicated for the other arm support.} 
The actuators are independently-controlled and operated via a custom-made pneumatic control board (Fig.~\ref{fig:pneu}) placed at a distance from the mannequin.\footnote{~The selected distance here was selected so as to match the one the board will be placed at during future testing with human subjects (e.g., at the back of a chair the infant will be seated at) for safety purposes.} 
The actuators are anchored on the mannequin via velcro straps at places that minimize motion restrictions as per our previous works~\cite{kokkoni2020development,mucchiani2022closed,sahin2022bidirectional}.\footnote{~Ongoing research aims to integrate the actuators onto a full wearable suit and study the effect of anchoring and placement on arm motion.} 

\begin{figure}[!t]
   \vspace{6pt}
   \centering
     \includegraphics[width=\columnwidth]{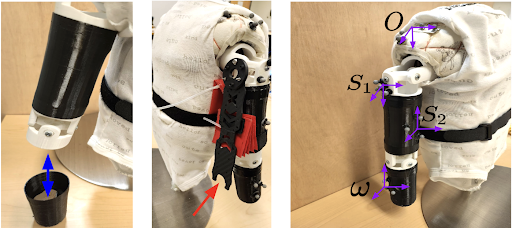}
      \vspace{-18pt}
      \caption{(Left) Our engineered mannequin built in-house integrates 3D-printed joints and links (upper-arm and forearm) designed based on anthropometric data. The links are adjustable (easily swappable and hollow to be filled in with additive weight) to make the mannequin easily reconfigurable. 
      (Center) Embedded supporting structure to mount one end of the shoulder flexion/extension actuator. (Right) The various coordinate systems employed in kinematics analysis and control of the overall system.}
      \label{fig:support}
      \vspace{-12pt}
   \end{figure}

The mannequin is designed and fabricated in-house based on average 12-month-old infant anthropometrics~\cite{Fryar2021,edmond2020normal}; the upper arm has a length of $l_a=0.14$\;m. The joints and upper-arm and forearm are 3D-printed. The upper-arm and forearm, in particular, are easily replaceable and hollow (left panel of Fig.~\ref{fig:support}).
We use a mix of different-sized beads to fill up the volume to achieve a desired weight so that density remains uniform and inertia/center of mass does not vary during operation. This design consideration allows us to rapidly vary the length and weight of the arm for testing different actuator designs and control parameters. In this work, the forearm of the mannequin is fixed at full extension so as to isolate and study shoulder joint motion alone.\footnote{~We refer the interested reader to~\cite{kokkoni2020development,mucchiani2022closed} for a study of the combined shoulder adduction/abduction and shoulder flexion/extension motion.}

Shoulder AB/AD motion (angle $\theta_1$) employs a recent design that has been extensively tested and validated in our previous work (Fig.~\ref{fig:act} a)~\cite{sahin2022bidirectional}, but a new actuator for shoulder F/E motion (angle $\theta_2$) was developed herein (Fig.~\ref{fig:act} b). 
Specifically, we consider a fabric-based bellow-type design actuator 
fabricated in a modular fashion, with interconnected square segments (pouches). Airflow through the inlet channel causes all pouches to simultaneously inflate; their total number determine the range of the actuator. 
Although proper characterization of the actuator is left as future work, based on empirical testing considering inflation time and range of extension, a total of $n=5$ pouches were selected herein. 
The soft pneumatic actuators employed herein can exert sufficient forces while maintaining a paper-thin profile. 
These are desirable characteristics when designing wearable devices that help minimize weight and impose less motion constraints. 
However, at the current design, F/E motion support via the bellow-type design requires a supporting element at one of its ends to allow the motion to occur (see the red arrow at the middle panel in Fig.~\ref{fig:support}). 


Two \textit{witmotion\footnote{~https://www.wit-motion.com/}} IMUs are attached at the upper-arm and forearm to estimate shoulder joint angles $\theta_1$ and $\theta_2$. A Nvidia Jetson Nano serves as the main computer for the overall system, running our developed controller and sending commands to the pneumatic board through two microcontrollers via serial communication. All the system runs on a Robot Operating System (ROS) framework we developed to combine the controller and sensor (IMU) readings along with actuation from the pneumatic control board.



\subsection{Desired Motion Characteristics}

\begin{figure}[!t]
\vspace{6pt}
\includegraphics[width=\linewidth]{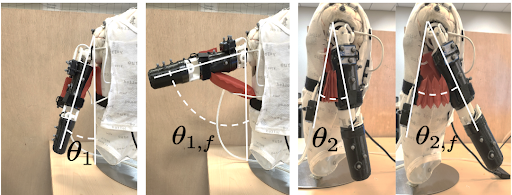}
\vspace{-18pt}
\caption{Range of motion afforded by the two actuators. (Left two panels) Shoulder adduction/abduction: $\theta_1\in[0.1745,1.3963]$\;rad; (Right two panels) Shoulder flexion/extension: $\theta_2\in[0.1745,0.5585]$\;rad.}
\label{fig:bellow}
\vspace{-6pt}
\end{figure}

The range of motion and associated kinematic constraints for the overall system are shown in Fig.~\ref{fig:bellow} and Table~\ref{table:constr}), respectively. At its current 5-pouch design, the bellow-type actuator assisting shoulder F/E motion can achieve a maximum angle variation of $\Delta\theta_2 = 0.3840 \; rad \;(22^o)$. 
While this is of relatively small F/E motion range, it does not affect testing the validity of our developed controller within the context of this work. In fact, the currently limited F/E motion range can be readily rectified by fabricating and employing actuators that integrate a large number of pouches, and is part of ongoing work. 
Fabricating different actuator designs of distinct length and number of pouches to extend or reduce the range of motion can be done quickly and in a cost-effective manner~\cite{sahin2022bidirectional}.
%
As far as the AB/AD range of motion is concerned, this was deemed satisfactory. 

\begin{table}[!ht]
\centering
\caption{Kinematic Constraints and Joint Velocity and Acceleration Limits at Initial ($t=0$) and Final ($t=T$) Time}
\label{table:constr}
\vspace{-6pt}
\begin{tabular}{ccccc}
\toprule
$\theta_i$ &  $\theta_{i,MIN}$ & $\theta_{i,MAX}$ &  
$\dot{\theta}_i$ \& $\ddot{\theta}_i$ ($t=0$) &
$\dot{\theta}_i$ \& $\ddot{\theta}_i$ ($t=T$) \\
\midrule
$1\;(F/E)$ & $0.1745$    & $1.3963$ & $0$  & $0$\\
$2\;(AB/AD)$ & $0.1745$    & $0.5585$ & $0$  & $0$\\
\bottomrule
\end{tabular}
\vspace{-8pt}
\end{table}

%
%
    

\subsection{Combined System Kinematics}

\begin{figure}[!t]
   \vspace{2pt}
   \centering
   \includegraphics[width=0.99\columnwidth]{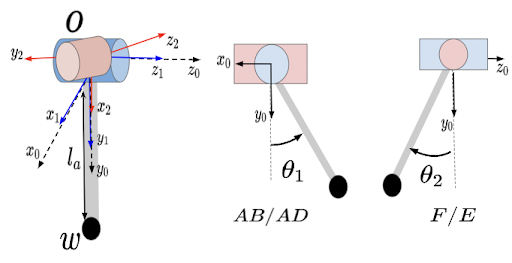}
      \vspace{-9pt}
      \caption{Coordinate frames for kinematic calculation of the abduction/adduction (AB/AD) and flexion/extension (F/E) of the shoulder joint.}
      \label{fig:coord}
      \vspace{-4pt}
\end{figure}

The last step prior to developing the controller is determination of the combined system (actuators and mannequin) kinematic model. We utilize the frames shown in the right panel of Fig.~\ref{fig:support}. Given that the elbow joint is taken to be fixed at full extension herein, we can consider a simplified model shown in Fig.~\ref{fig:coord} and readily derive its Denavit-Hartenberg (DH) parameters as given in Table~\ref{table:dh}.

\begin{table}[!ht]
\vspace{0pt}
\centering
\caption{Denavit-Hartenberg Parameters}
\label{table:dh}
\vspace{-6pt}
\begin{tabular}{cccc}
\toprule
\textbf{$\theta_j$} & \textbf{$d_j$} & \textbf{$r_j$}  & \textbf{$\alpha_j$} \\ \midrule
$\theta_1$ & $0$    & $0$ & $\pi/2$                     \\
$\theta_2$ & $0$    & $l_a$ & $\pi/2$ \\
\bottomrule
\end{tabular}
\vspace{-12pt}
\end{table}
 
The transformation matrix between the wrist ($w$) and the shoulder origin ($o$) is 

\begin{gather}
  \scalebox{1}{%
  $T =$
        $  \begin{bmatrix}
   c\theta_1c\theta_2 &   -c\theta_1 s \theta_2    & -s\theta_1 & l_ac\theta_1c\theta_2 \\
   c\theta_2s\theta_1 &  -s\theta_1 s \theta_2 &   c\theta_1   & l_ac\theta_2s\theta_1 \\
   -s\theta_2  & - c\theta_2  & 0  & -l_a s\theta_2  \\
      0 & 0 & 0 &  1\\
  \end{bmatrix},
$
    \label{eq:forward}}
  \end{gather}
where the Cartesian position of the wrist $w=[x,y,z]^T$ is
 \begin{gather}
  \scalebox{1}{%
  $\begin{bmatrix}
   x \\
    y\\
    z\\
  \end{bmatrix}=$
   $\begin{bmatrix}
   l_ac\theta_1c\theta_2 \\
    l_ac\theta_2s\theta_1\\
   - l_a s\theta_2\\
  \end{bmatrix}\enspace.
  $
   \label{eq:forw}}
 \end{gather}
For brevity, we use shorthands $c\cdot \equiv cos(\cdot)$ and $s\cdot \equiv sin(\cdot)$ in~\eqref{eq:forward}--\eqref{eq:forw}. 
Then, the respective shoulder angles for AB/AD ($\theta_1$) and F/E ($\theta_2$) can be calculated as 
%
$\theta_1 = atan2\left(\frac{y}{x}\right)$, and 
$\theta_2 = asin\left( -\frac{z}{l_a} \right)$. 

\section{Low-level Trajectory Tracking Control}
\label{sec:methods}

\subsection{Overview}
Given a desired reference trajectory in joint space position, velocity and acceleration  $\theta_d(t),\dot{\theta}_d(t)$ and $\ddot{\theta}_d(t)$, let the tracking and input errors be $e = \theta - \theta_d(t)$ and $e_u = u - u_d(t)$, respectively. Our closed-loop control system then is
\begin{equation}
\begin{aligned}
        u(t) = u_d(t) - K_{GPI}(\theta-\theta_d(t))\enspace, \\
        \ddot{\theta}(t) = u(t) + \rho(t)\enspace,
\end{aligned}
    \label{eq:ccon1}
\end{equation}
\noindent where $K_{GPI}$ is the proposed controller, $\rho(t)$ a polynomial of degree $r$ describing disturbances to the system, and the nominal control input $u_d(t)$ can be determined according to the unperturbed system ($\rho(t)=0$). The proposed GPI controller can be described as a lead compensator, i.e.

\begin{equation}
\begin{aligned}
K_{GPI}(s) &= \left(\frac{k_{r+2,i}s^{r+2}+k_{r+1,i}s^{r+1}+...+k_{1,i}s + k_{0,i}}{s^{r+1}(s+k_{r+3,i})} \right)\enspace,\\
    \end{aligned}
    \label{eq:each}
\end{equation}
with controller gains $k_{j,i}$; indices $j=0,\ldots,r+2$ and $i=1,2$ denote the controller constant and actuator, respectively. 
Recall that each actuator is controlled independently, hence~\eqref{eq:each} applies independently for $i=1,2$.\footnote{~To improve clarity, indices denoting actuator number are dropped when presenting functional expressions that apply to both actuators (e.g., $K_{GPI}$).}

\begin{figure}[!t]
    \vspace{3pt}
   \centering
     \includegraphics[trim={0.5cm 0 0.5cm 0},clip,width=\columnwidth]{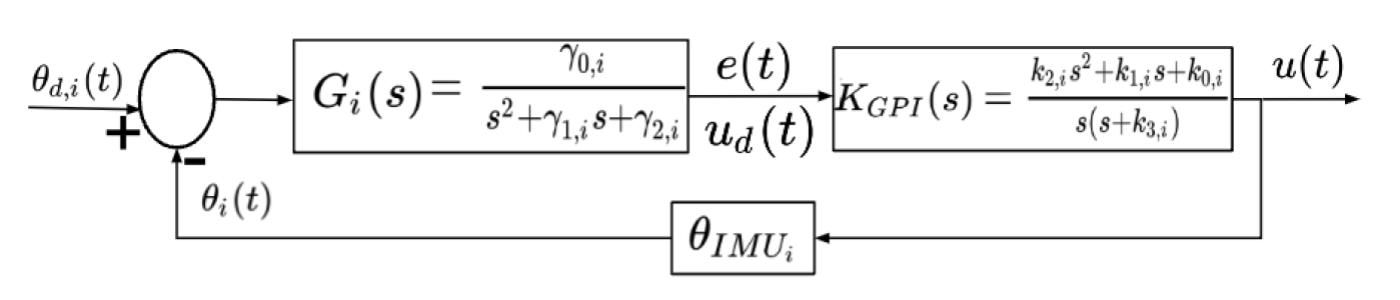}
      \vspace{-16pt}
      \caption{Schematic of the proposed GPI controller for each actuator, $i=1,2$. Controller parameters are tuned experimentally. The mannequin's joint angle information is provided in real-time via proprioceptive feedback.}
      \label{fig:cl}
      \vspace{-6pt}
\end{figure}

The control loop is depicted in Fig.~\ref{fig:cl}. For each DoF, estimated transfer function parameters $[\gamma_{0i},\gamma_{1i},\gamma_{2i}]$ and parameters $\xi_i$ and $\omega_{ni}$ need to be set. The controller acts on the error between the desired joint values and actual ones provided via IMU readings. 
In the following, we present the controller derivation and how its parameters are tuned. 

\subsection{Controller Derivation}
In this work we assume $r=0$ (i.e. a constant disturbance).  Then~\eqref{eq:each} reduces to
\begin{equation}
    K_{GPI}(s) = \frac{k_{2,i}s^2+k_{1,i}s+k_{0,i}}{s(s+k_{3,i})}\enspace.
\end{equation}
Treating each DoF of the combined mannequin and actuators system as a spring-mass-damper system, the transfer function which relates the PWM \% of the pneumatic actuator and angle $\theta_i$ 
attains the form
\begin{equation}
    G_i(s) = \frac{\gamma_{0,i}}{s^2 + \gamma_{1,i} s + \gamma_{2,i}}\enspace.
    \label{eq:mass}
\end{equation}
 
\noindent Rewriting~\eqref{eq:ccon1}, and dividing with $\gamma_{0_i}$ yields 
\begin{equation}
        u(t) = u_d(t) - \left(\frac{1}{\gamma_{0,i}}\frac{k_{2,i}s^2+k_{1,i}s+k_{0,i}}{s(s+k_{3,i})}\right)e
        \label{eq:imple}
\end{equation}
since $e(s)=G_i(s)e_u(s)$ relates the input and output errors. Combining with~\eqref{eq:mass}, we get
\begin{equation}
    \left[1+\left( \frac{1}{s^2 + \gamma_{1,i} s + \gamma_{2,i}} \right) \left( \frac{k_{2,i}s^2+k_{1,i}s+k_{0,i}}{s(s+k_{3,i})}\right)\right]e=0
    \label{eq:both}
\end{equation}
leading to the characteristic equation 
\begin{equation}
\begin{aligned}
    s^4 + (k_{3,i}+\gamma_{1,i})s^3 + (k_{2,i} + k_{3,i}\gamma_{1,i}+\gamma_{2,i})s^2 +\\ (k_{3,i}\gamma_{2,i}+k_{1,i})s + k_{0,i} = 0\enspace.
    \label{eq:1}
    \end{aligned}
\end{equation}

The controller gains $k_{j,i}$ can be determined by factoring the Hurwitz polynomial 
%
    $(s^2 + 2\xi w_n^2+w_n^2)^2 = 0
$, 
%
which yields
\begin{equation}
    \begin{aligned}
        &k_{0,i} = w_n^4\enspace, \\
        &k_{1,i} = 4w_n^3\xi - \gamma_{2,i}(4\xi w_n-\gamma_{1,i})\enspace,\\
        &k_{2,i} = 2w_n^2 + 4\xi^2w_n^2 - \gamma_{1,i}(4\xi w_n - \gamma_{1,i}) - \gamma_{2,i}\enspace,\\
        &k_{3,i} = 4\xi w_n - \gamma_{1,i}\enspace.
    \end{aligned}
    \label{eq:third}
\end{equation}

Therefore, our controller can be determined with parameters from~\eqref{eq:third}. The controller design parameters are $\xi_i$ and $w_{n_i}$. From~\eqref{eq:imple} we can implement the controller as 
\begin{equation}
\begin{aligned}
    u = u_d - k_{3,i}(\dot{\theta}_{int}-\dot{\theta}_d) + \\ \frac{1}{\gamma_{0,i}}\left[ -k_{2,i}(e-e_0) - k_{1,i}\int_t e dt  - k_{0,i}\iint_t e dt \right] \enspace,
    \end{aligned}
    \label{eq:controlad}
\end{equation}
%
where $\theta_{int}$ is the integral reconstruction \cite{romero2014algebraic,blanco2022robust} of input $u$

\begin{equation}
        \theta_{int} = \int_{0}^{t} u dt~,~~~~
        \dot{\theta} = \theta_{int} -\dot{\theta}_0
            \label{eq:ust}
\end{equation}
and the value $u_d$ is given by~\cite{sira2007fast}, 
\begin{equation}
    u_d = \frac{1}{\gamma_{0,i}}[\ddot{\theta}_d + \gamma_{1,i}\dot{\theta}_d+\gamma_{2,i}\theta_d]\enspace.
\end{equation}
For both~\eqref{eq:controlad} and~\eqref{eq:ust}, the integral values are numerically estimated by the trapezoidal rule. 

\subsection{Parameter Tuning}

\begin{figure}[!t]
   \vspace{5pt}
   \centering
     \includegraphics[width=0.9\columnwidth]{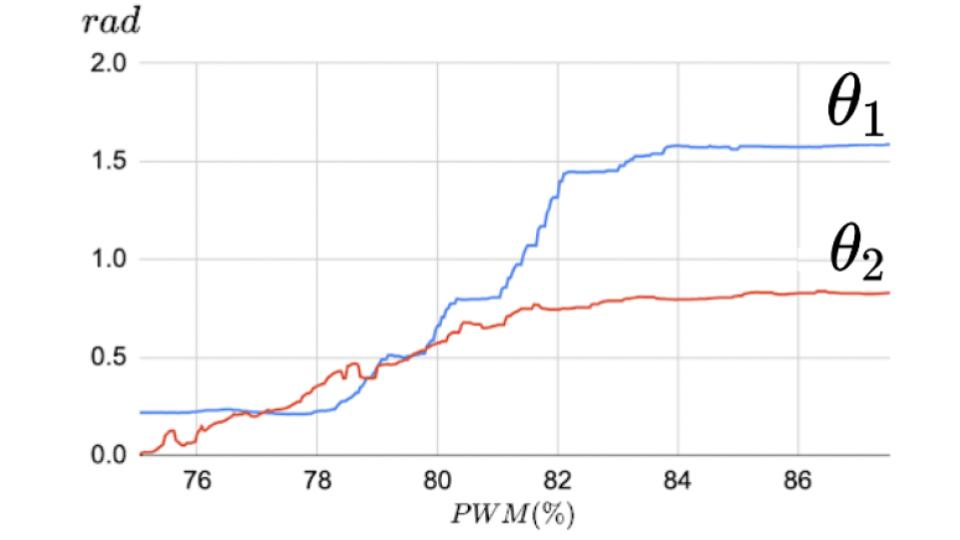}
      \vspace{-9pt}
      \caption{Empirical correlation of PWM \% and joint angle variations. 
      }
      \label{fig:datap}
      \vspace{-9pt}
\end{figure}

Each $G_i(s)$ is estimated from offline experimental data and is inputted to Matlab ($tfest$ function with sampling time of $T_s=0.065s$, 700 data-points) to compute variations of both DoF of the shoulder and PWM percentages of the pneumatic pumps (shown in Fig.~\ref{fig:datap}) leading to
   \begin{equation}
   \begin{aligned}
     G_1 &= \frac{0.0005725}{s^2+0.05725s+0.044}\enspace, \\
    G_2 &= \frac{0.0003665}{s^2+0.213s+0.04079}\enspace.
   \label{eq:tf}
   \end{aligned}
 \end{equation}

$G_1$ and $G_2$ relate to shoulder AB/AD (estimation fit of $89\%$) and F/E (estimation fit of $91.36\%$) motion, respectively. We set $\xi_1$, $\xi_2 = 0.9$, $\omega_{n,1}=6.1$ \;rad/s and $\omega_{n,2}=10.25$ \;rad/s. These parameters are essential to compute controller gains as per~\eqref{eq:third}, and were chosen so that the input $u(t)$ is identically set at a maximum value of $100$, i.e. $100\%$ PWM.

\section{Experimental Results}
\label{sec:experim}

\begin{figure}[!t]
\vspace{3pt}
\includegraphics[width=\linewidth]{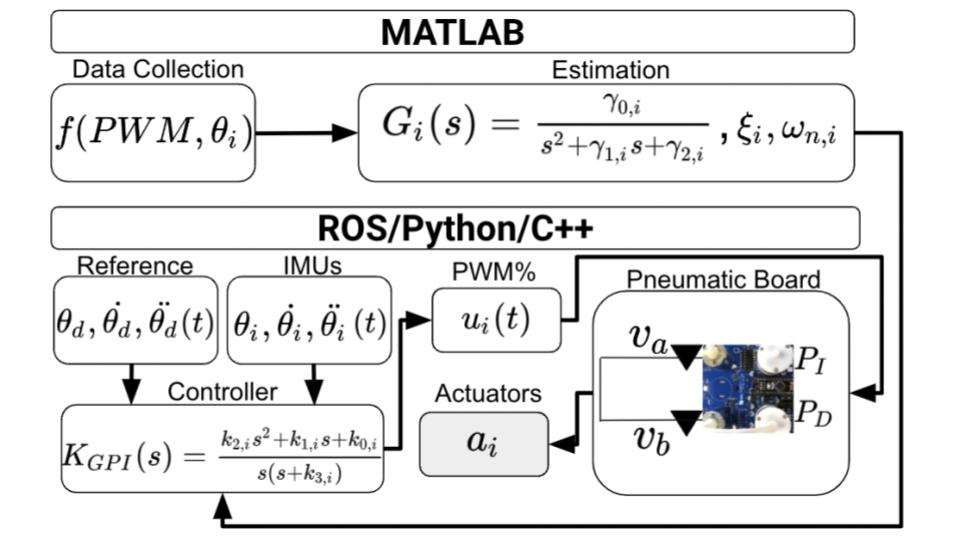}
\vspace{-21pt}
\caption{Overall structure of the system implementation.}
\label{fig:overall}
\vspace{-3pt}
\end{figure}

The overall system flow diagram is shown in Fig.~\ref{fig:overall}. 
We test three types of trajectories. The first set includes interpolated trajectories (via quintic polynomials) given desired end-points for both AB/AD and F/E shoulder motion separately, as well as simultaneously. The second set considers periodic trajectories in the form of harmonic signals while the third set focuses on teach-and-repeat trajectories. 

\subsection{End-Point Trajectories}

To support motion smoothness, intermediary desired joint angle values are computed based on quintic polynomial time scaling, $\theta_d(t) = a_0t^5+a_1t^4+a_2t^3+a_3t^2+a_4t+a_5,~t\in[0,T]$, for each joint separately. The polynomial coefficients are calculated based on the kinematic constraints shown in Table~\ref{table:constr} for full-range motion. For distinct end-point trajectories considered during this part of experimentation, the initial ($t=0$) and terminal ($t=T$) time joint velocities and accelerations are all set to 0, while the initial and final joint angles may vary (Table~\ref{table:trajec}). 

\begin{table}[!t]
\vspace{0pt}
\centering
\caption{Desired Trajectory End-Points}
\label{table:trajec}
\vspace{-6pt}
\begin{tabular}{ccccc}
\toprule
Case & $\theta_1 [rad]$ & $\theta_2 [rad]$ & $(x,y,z) [m]$ & Type\\ \midrule
$q_1$ & $0.6981$ & $-$    & $[0.107,0.107,0]$     & $AB/AD$              \\

$q_2$ & $1.0472$ & $-$    & $[0.070,0.070,0]$    & $AB/AD$                \\

$q_3$ & $-$ & $ 0.3491$   & $[0.132,0.132, -0.048]$   & $F/E$                \\

$q_4$ & $-$ & $0.5585$    & $[0.119,0.119, -0.074]$             & $F/E$       \\

$q_5$ & $0.6981$ & $0.3491$    & $[0.066,0.066, -0.048]$      & $AB/AD$ \& $F/E$             \\

$q_6$ & $0.6981$ & $0.5585$    & $[0.059, 0.059, -0.074]$      & $AB/AD$ \& $F/E$             \\

$q_7$ & $ 1.3963$ & $0.3491$    & $[0.023, 0.023,-0.048]$      & $AB/AD$ \& $F/E$             \\

$q_8$ & $1.3963$ & $0.5585$    & $[0.021,0.021,-0.074]$         & $AB/AD$ \& $F/E$        \\
\bottomrule
\end{tabular}
\vspace{-15pt}
\end{table}











     \begin{figure*}[!t]
     \vspace{6pt}
   \centering
     \includegraphics[width=0.80\textwidth]{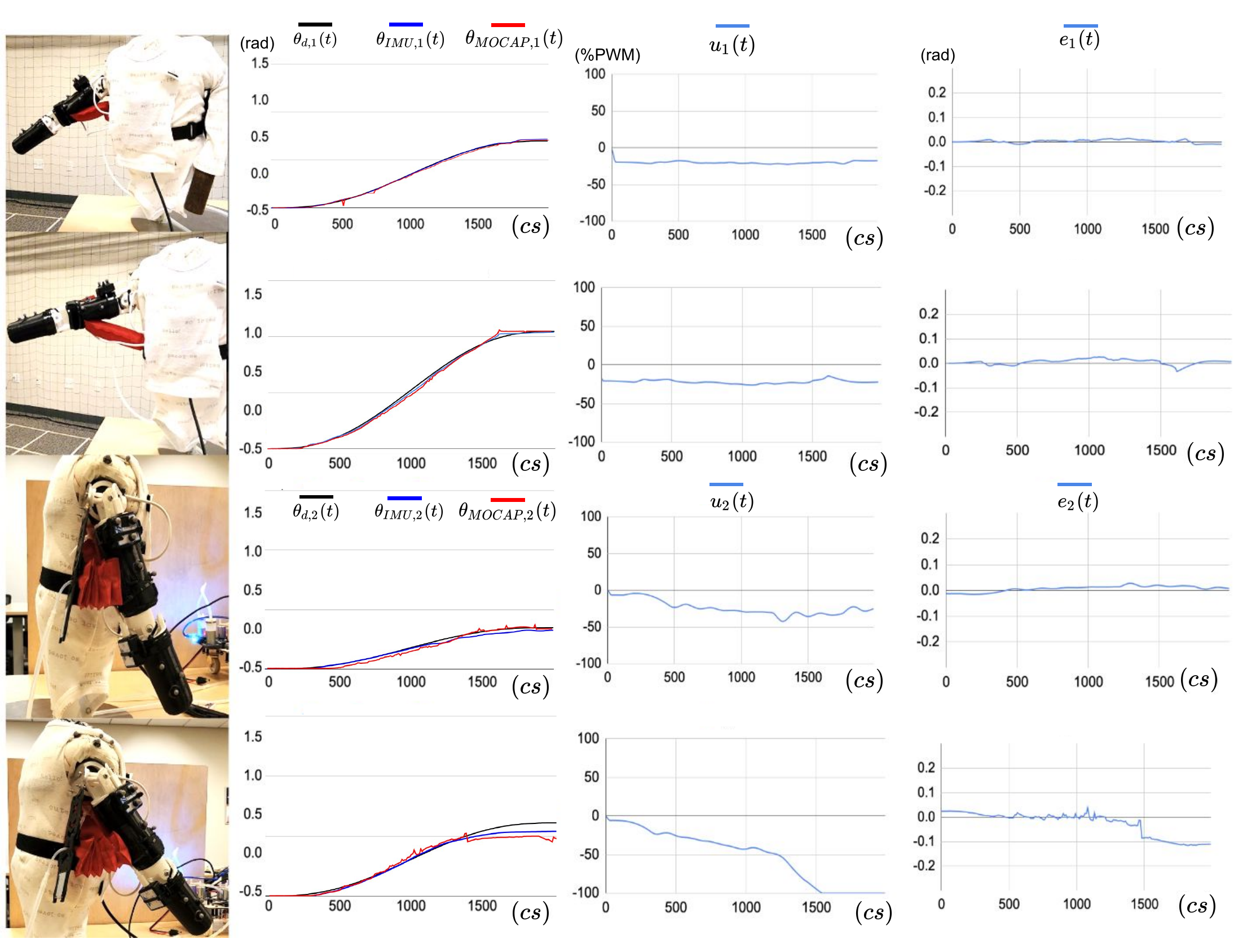}
      \vspace{-12pt}
      \caption{Experimental results for \emph{single actuator operation} cases $q_1$ through $q_4$ (from top to bottom). (Left) Averaged evolution of angle (desired, actual via IMU, and ground truth via motion capture). (Center) Averaged control input evolution. (Right) Averaged error evolution. Individual trials in each case were very close to each other, and hence for clarity of presentation only the mean values are presented. Time is shown in centiseconds ($10^{-2}s$).}
      \label{fig:gf1}
      \vspace{-3pt}
   \end{figure*}

       \begin{figure*}[!ht]
       \vspace{-6pt}
   \centering
     \includegraphics[width=0.80\textwidth]{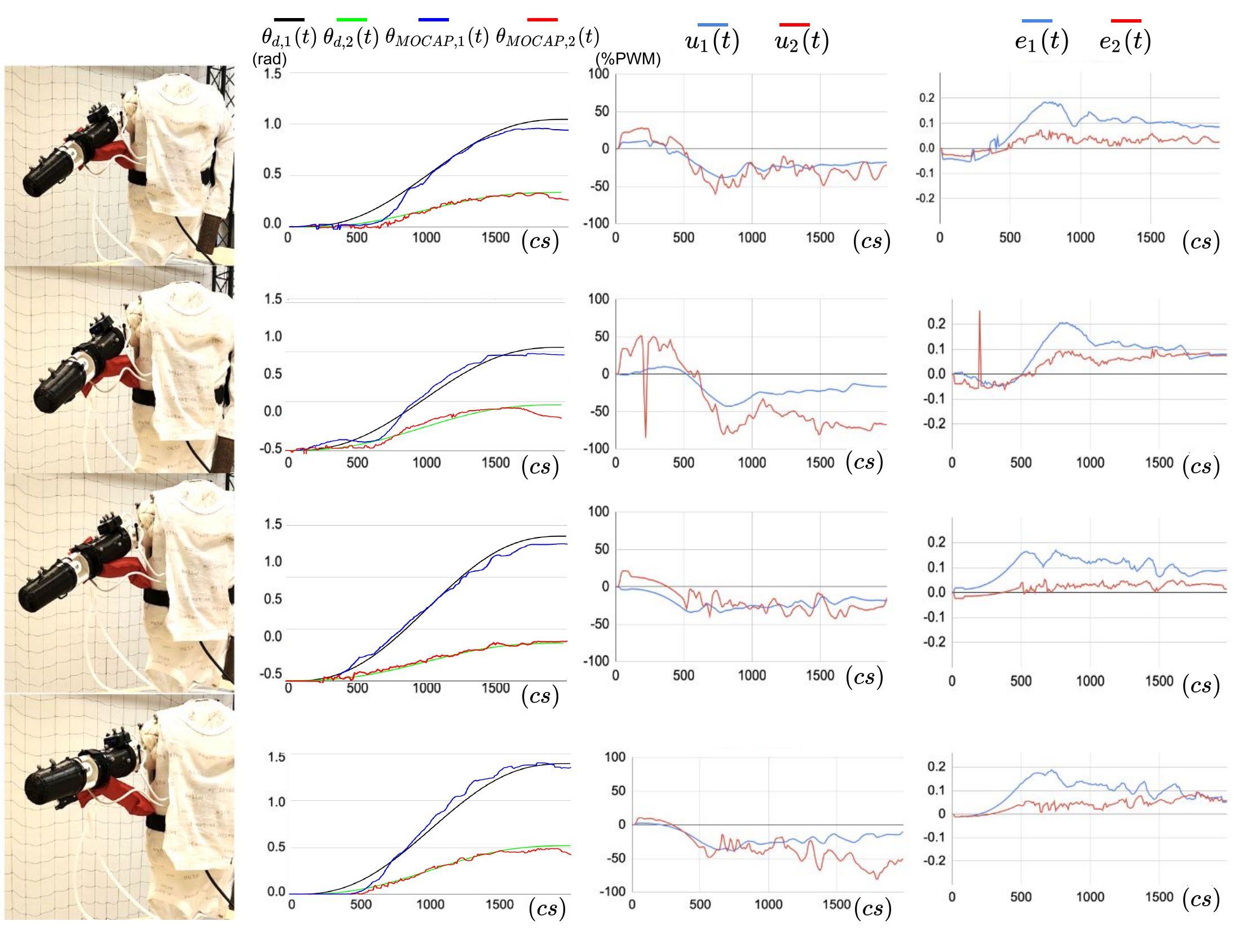}
      \vspace{-18pt}
      \caption{Experimental results for \emph{simultaneous actuator operation} cases $q_5$ through $q_8$ (from top to bottom). 
      (Left) Averaged evolution of angle (desired, actual via IMU, and ground truth via motion capture). (Center) Averaged control input evolution. (Right) Averaged error evolution. Individual trials in each case were very close to each other, and hence for clarity of presentation only the mean values are presented. Time is shown in centiseconds ($10^{-2}s$).
      }
      \label{fig:gf2}
      \vspace{-6pt}
   \end{figure*}

We performed a total of 60 experimental trials considering the various interpolated trajectories from the end-points listed in Table~\ref{table:trajec} (seven trials for cases $q_1$ through $q_4$ each, and eight trials for each of the remainder cases). The starting angle was set at $\theta_i=0.2$\;rad as per kinematic constraints for all cases.  %
Figures~\ref{fig:gf1} and~\ref{fig:gf2} summarize the obtained results (average values depicted). 
Actual (integrated) IMU feedback was compared against ground truth measurements provided by a 12-camera Optitrack motion capture system. 
%
Steady-state errors (mean squared errors [MSE] and standard deviation errors [SDE]) are shown in Table~
\ref{table:mse_1} for the single actuator operation cases $\{q_1,q_2,q_3,q_4\}$, and in Table~\ref{table:mse_2} for the simultaneous actuator operation cases $\{q_5,q_6,q_7,q_8\}$.

\begin{figure*}[!t]
\vspace{6pt}
\centering
\includegraphics[width=\textwidth]{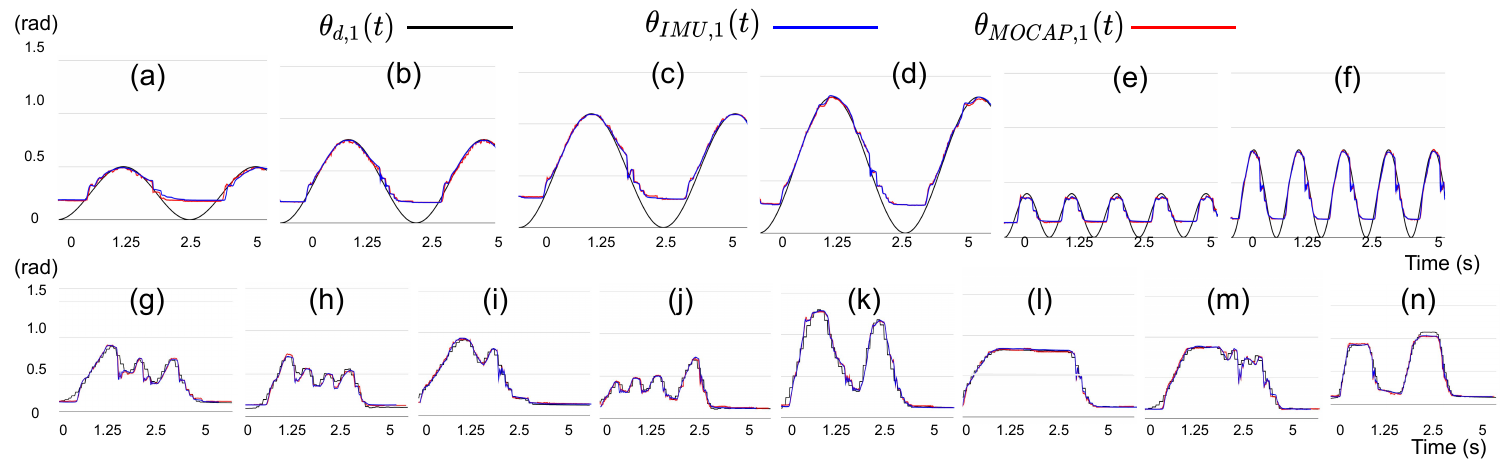}
\vspace{-21pt}
\caption{Top: Experiments with various harmonic trajectories ($h=300$) and: a ($A=1,\;f=1.6\;e^{-3}$), b ($A=1,\;f=2.6 \; e^{-3}$), c ($A=1,\;f=3.6\;e^{-3}$), d ($A=1,\;f=4.3\;e^{-3}$), e ($A=3,\;f=1.3\;e^{-3}$) and f ($A=3,\;f=2.6\;e^{-3}$). Bottom: Custom teach-and-repeat trajectories. All curves denote averaged quantities. Repeatability was high and hence for clarity standard deviations funnels are not shown.}
\label{fig:siteach}
\vspace{-9pt}
\end{figure*}

Overall, it can be readily verified that our proposed controller can successfully track the desired trajectories smoothly and with minimal steady-state error. In single-actuator operation (Fig.~\ref{fig:gf1}) all cases but one ($q_4$) perform as desired. In this case, the desired end-point is at the limit $\theta_2=0.5585$\;rad, and the control input signal $u_i(t)$ (depicted in the middle columns) saturates (i.e. 100\% PWM signal) when attempting to inflate the actuator. This means the controller could not overcome the physical limitation of the actuator while acting against the total weight of the mannequin's arm, and so it was not able to achieve the desired end-point. Additionally, relative motion of the back support with respect to the arm itself was observed, leading to a higher steady state error. However, no limitations were observed for trajectories within the actuator's physical boundary, as shown in the tracking performance for end-point $ \theta_2=0.3491$\;rad. In the combined trajectories (cases $q_5$ through $q_8$) we did not observe any control saturation even when $\theta_2$ was commanded to the limit. This can be associated to implicit actuator synergies whereby one may aid another at different levels of pressurization. The tracking errors are still small (cf MSE in Tables~\ref{table:mse_1} and~\ref{table:mse_2}) but relatively larger compared to those in single-actuator operation. Comparably increased errors can be associated with higher motion variability attained due to no explicit coupling of the two actuators. 



\begin{table}[!t]
\vspace{6pt}
\caption{MSE (SDE) for Joint Angles $\theta_1$ and $\theta_2$ during Single Actuator Operation}
\vspace{-6pt}
\resizebox{\columnwidth}{!}{%
\begin{tabular}{@{}cllll@{}}
\toprule
 & $q_1$ & $q_2$ & $q_3$ & $q_4$  \\ \midrule
$e_{\theta_1}10^{-3}$ rad & 0.04 (0.20) & 0.08 (0.28) &  & \\
$e_{\theta_2}10^{-3}$ rad &  &  & 0.06 (0.24) & 1.00 (1.00) \\ 
\bottomrule
\end{tabular}%
}
\label{table:mse_1}
\vspace{-6pt}
\end{table}

\begin{table}[!t]
\vspace{4pt}
\caption{MSE (SDE) for Joint Angles $\theta_1$ and $\theta_2$ during Simultaneous Actuator Operation}
\vspace{-6pt}
\resizebox{\columnwidth}{!}{%
\begin{tabular}{@{}cllll@{}}
\toprule
 &$q_5$ & $q_6$ & $q_7$ & $q_8$ \\ 
 \midrule
$e_{\theta_1}10^{-3}$ rad & 1.00 (1.00) & 1.00 (1.00) & 1.00 (1.00) & 1.00 (1.00) \\
$e_{\theta_2}10^{-3}$ rad & 1.00 (1.00) & 4.00 (2.00) & 0.70 (0.83) & 2.00 (1.41) \\ \bottomrule
\end{tabular}%
}
\label{table:mse_2}
\vspace{-9pt}
\end{table}

\subsection{Harmonic Trajectories}
To test our method's response to periodic motion, harmonic trajectories of various frequency and amplitudes were evaluated with the AB/AD actuator (given its wider range of motion). Desired trajectories were as  
%
       $\theta_1(t) = A/2sin(ft+h) + A/2$ 
%
($A$: amplitude, $f$: frequency, $h$: constant). 

Results for various trajectories are shown in Fig.~\ref{fig:siteach} (a to f), with a total of 25 experimental trials. It can be seen that the device correctly follows the desired trajectories, while small deviations occurring on valleys (seen as straight lines). This can be associated to physical constraints on the minimum achievable AB/AD joint angle (Fig.~\ref{fig:bellow}), thus not allowing the actuator to reach a closer to $0$\;rad value. 

\subsection{Teach-and-Repeat Trajectories}
To investigate more versatile trajectories, particularly aiming at evaluating the controller within the achievable range of the AB/AD actuator motion, we implemented a teach-and-repeat mode on the wearable. Desired trajectories were given for five seconds by a researcher physically manipulating the mannequin, and the controller was task to repeat it afterwards. To record the demonstrated trajectories, input from the IMU angular displacement and velocity were directly recorded, while acceleration was estimated by differentiation.

Results are shown in Fig.~\ref{fig:siteach} (g to n). We performed 25 experimental trials in this mode. The controller is seen to closely follow all taught trajectories, with minor deviations attributed to vibrational noise on the IMU, introduced by the higher frequency segments (as seen in Fig. \ref{fig:siteach} cases g, h and j). The accompanying video demonstrates these experiments.


\section{Conclusion}
\label{sec:conclusion}

This work introduced a closed-loop control method based on proprioceptive feedback for pressure regulation of soft pneumatic actuators embedded on an infant-scale engineered mannequin so as to follow desired trajectories in support of shoulder motion. The developed low-level controller can help track higher-level force-control-based trajectories in future pediatric wearable robotic assistive devices. 
%
Extensive experimentation confirmed the ability of our proposed controller to track a range of diverse trajectories (desired end-points, harmonic, and teach-and-repeat) smoothly and accurately.

This work conducted herein lays the basis for several future research directions. Notably, we aim to integrate shoulder and arm support. Further we seek to embed the actuators onto a complete wearable device and further test it with the engineered mannequin. Once the efficacy and safety of the wearable is demonstrated, we plan on moving with human subjects testing. 


\bibliographystyle{ieeetr}
\bibliography{root}

\end{document}